\documentclass{article}

     \PassOptionsToPackage{numbers, compress}{natbib}

 \usepackage[preprint]{neurips_2021}

\usepackage[utf8]{inputenc} 
\usepackage[T1]{fontenc}    
\usepackage{hyperref}       
\usepackage{url}            
\usepackage{booktabs}       
\usepackage{amsfonts}       
\usepackage{nicefrac}       
\usepackage{microtype}      
\usepackage{xcolor}         
\usepackage{tikz}
\usepackage{caption}
\usepackage{subcaption}
\usepackage{makecell}

\title{Armour: Generalizable Compact Self-Attention for Vision Transformers}

\author{%
  Lingchuan Meng\\
  Arm Inc.\\
  San Jose, CA 95134 \\
  \texttt{lingchuan.meng@arm.com} \\
}

\begin{document}

\maketitle

\begin{abstract}
Attention-based transformer networks have demonstrated promising potential as their applications extend from natural language processing to vision.  However,  despite the recent improvements, such as sub-quadratic attention approximation and various training enhancements, the compact vision transformers to date using the regular attention still fall short in comparison with its convnet counterparts, in terms of \textit{accuracy,}  \textit{model size},  \textit{and} \textit{throughput}.  This paper introduces a compact self-attention mechanism that is fundamental and highly generalizable.  The proposed method reduces redundancy and improves efficiency on top of the existing attention optimizations.  We show its drop-in applicability for both the regular attention mechanism and some most recent variants in vision transformers.  As a result, we produced smaller and faster models with the same or better accuracies. 
\end{abstract}

\section{Introduction }
\label{sec:introduction}
The Transformer architecture  \citep{transformer} has been widely used in natural language processing (NLP) tasks and obtained dominating results.  Recently,  the Vision Transformer (ViT) \citep{vit} has demonstrated the state-of-the-art results for image classification by using exclusively the Transform architecture and the multi-head attention (MHA).  However,  as a proof-of-concept, ViT underperforms the convolutional neural networks (CNN) in accuracy unless a large private labeled image dataset (JFT-300M) is used.  DeiT \citep{deit} first introduced some smaller variants of ViT by reducing the embedding dimension and the number of heads.  Unfortunately,  the smallest DeiT architecture still falls short when compared against the CNN counterpart,  EfficientNet B0 \citep{efficientnet},  across all the key metrics (\textit{accuracy},  \textit{model size,} and \textit{latency}.) Since then,  several works \citep{cvt, levit} proposed hybrid architectures that reintroduce convolutions alongside the MHAs, as the desirable inductive biases built into the former are uniquely suited to solve vision tasks.

In this paper, we re-examine the fundamental design of self-attention for opportunities to further reduce model size and latency for vision transformers, while at least maintaining the same accuracy.  We are especially interested in generalizable optimizations that can be easily applied on top of both the widely-used regular self-attention and its latest variants. 

To achieve the goal,  we first identify the redundancy in the linear transformations in self-attention that produce the query,  key,  value matrices.  Then we propose a compact self-attention called \textit{Armour} which reduces the number of transformations,  therefore improving both model size and latency.  The proposed method is compared against its variants, and shows preferable performance in accuracy and throughput.  Next, we replace the regular self-attention with Armour in the smallest accurate ViT in \citep{deit},  which yields a new model with the same accuracy,  while achieving EfficientNet B0-level model size and throughput.  

Since Armour targets the fundamental design redundancy of self-attention,  it is highly \textit{generalizable} and can be applied on top of most ViT optimizations.  To demonstrate its drop-in applicability,  we replace the evolved attention layers in a competitive hybrid ViT architecture \citep{levit} with two Armour variants. The resulting models achieve better accuracy and GPU/CPU throughputs with fewer parameters.  Given the all-round advantages, the easy-to-implement Armour attention and its generalized forms should be considered in places where self-attention and its variants are used,  especially for the rapidly evolving vision transformers. 

\section{Related Work}
\label{sec:related_work}
The attention mechanism was first introduced and used in conjunction with a recurrent network in \citep{nmt} to help memorize long sequences in neural machine translation.  The Transformer \citep{transformer} further proposed the multi-head attention mechanism which performs multiple self-attention in parallel to achieve more effective learning at different scales.  Recently,  the Transformer architecture has extended its application from NLP to vision tasks,  especially as ViT \citep{vit} has achieved the state-of-the-art results for image classification with an architecture that closely resembles the initial NLP version.  The other examples of using the Transformer as a viable alternative to CNNs include object detection \citep{detr, up_detr, DBLP:journals/corr/abs-2011-10881, DBLP:journals/corr/abs-2010-04159}, video processing \citep{DBLP:journals/corr/abs-2007-10247, DBLP:journals/corr/abs-1804-00819}, image enhancement \citep{DBLP:journals/corr/abs-2012-00364, DBLP:journals/corr/abs-2006-04139}, etc.  Due to the enormous computational and memory requirements of the transformer models,  there has been increasing interest in transformer optimizations along several directions. 

The regular transformer scales quadratically with the number of tokens $N$ in the input sequence, which limits its usage in settings with limited resources.  Several works \citep{linformer, reformer,  performer, transformer_rnn} have proposed techniques to reduce the complexity from $O(N^2)$ to as low as $O(N)$,  yielding linear transformers.  However, these techniques only outperform the regular method for very long sequences. 

Another direction focuses on the sparsity in transformers. \citep{sparse_transformer} exploits a factorized sparse representation of attention. \citep{spatten} leverages token and head sparsities to reduce the attention computation and memory access for NLP tasks.  For ViT, \citep{vit_pruning} is an early example that prunes channels according to the learnable importance scores.  In our work, the reduction of the transformations in self-attention can be viewed as a fixed layer pruning scheme, which is unexplored by any of these works.

Evolving the Transformer architecture can also significantly improve  performance. For ViTs, the evolutions can be categorized by how the key components are implemented,  including positional encoding\citep{levit, cvt}, token embedding \citep{t2t}, projection for attention \citep{pvt}, and model hierarchy \citep{pvt, cvt, levit}.  Selecting deployment-friendly operators for non-linearity and normalization can also greatly improve inference latency. For example,  \citep{levit} replaced the linear projection and LayerNorm \citep{layernorm} with 1$\times$1 convolutions and BatchNorm\citep{batchnorm}, respectively,  later of which can be \textit{fused} into the preceding convolution at inference time.  Our work introduces optimizations at the fundamental level of the self-attention mechanism.  As a result, it is high generalizable and can be applied on top of most of the aforementioned techniques to achieve additional improvements in accuracy, model size, and throughput. 

\section{Compact Self-Attention}
\label{sec:compact_self_attention}
In this section,  we first review the preliminaries associated with the regular attention mechanism, and raise the question of whether it should be adopted verbatim in the context of non auto-regressive tasks, such as the vision transformer.  Then, we identify the redundancy in the linear transformations and propose the Armour attention to exploit it.  

\subsection{Preliminaries}
The \textit{scaled dot-product attention} is a mapping that accepts three matrices representing the packed vectors of \textit{query},  \textit{key} and \textit{value},  respectively.  Conventionally,  these matrices are named as $Q,  K,  V \in \mathbb{R}^{L\times d}$,  where $L$ is the length of the input token sequence,  and $d$ is the hidden dimension.  Then scaled dot-product attention can be defined as:
\begin{equation}
\label{eqn:dot_prod_atten}
\textrm{Attention}(Q,  K,  V) = softmax(QK^T / \sqrt{d}) V,
\end{equation}
where the \textit{softmax} function is applied row-wise for each token to compute the attention probabilities.  When the $Q$, $K$, and $V$ matrices are computed from the same input tensor $x$,  i.e.,  $Q=xW_Q, K=xW_K, V=xW_V$,  it is regarded as a \textit{self-attention}.  Finally,  a \textit{multi-head attention} (MHA) performs $h$ self-attentions in parallel over the attention heads,  which collectively allow more effective learning at different scales. 

The attention mechanism can be interpreted as a retrieval system in matrix form.  The dot-product between $Q$ and $K^T$ calculates the similarities between the query and key vectors, known as the attention scores.  Then, the scores are normalized by the softmax function to produce the attention probabilities which indicate how much each token will be expressed at a given position.  Finally, the probability matrix is multiplied with $V$ to yield weighted value vectors which are summed up as the output for each position.  Originally, the attention mechanism has been used mostly in auto-regressive models for NLP tasks, meaning that $Q$ is not calculated from the same sequence as $K$ and $V$.  But for self-attention in ViTs, all three matrices are linearly transformed from the same input sequence.  \textit{Do we still need all three linear transformations?}

\subsection{Armour Attention}
\begin{figure}
  	\centering
  	\includegraphics[width=.95\textwidth]{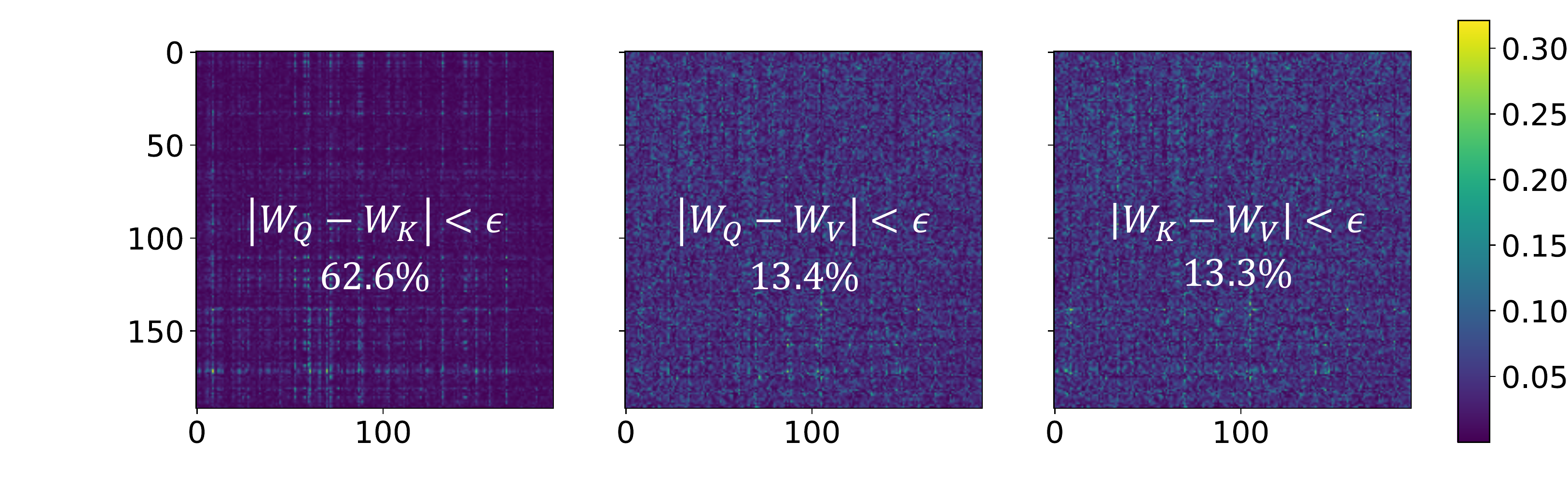}
  	\caption{Redundancies between $W_Q$, $W_K$, and $W_V$ in the regular self-attention.}
  	\label{fig:grad_entangling}
\end{figure}

We take a closer look at the linear transformations.  Recall that $Q=xW_Q$ and $K=xW_K$ in $QK^T$.  Since $Q$ and $K$ are linearly transformed from the same input $x$,  the weight matrices $W_Q$ and $W_K$ are \textit{entangled in gradient back-propagation. } This is a fundamental redundancy in the regular self-attention mechanism.  Figure \ref{fig:grad_entangling} visualizes the pair-wise redundancies between $W_Q, W_K,$ and $W_V$ in the first attention layer of DeiT-Ti \citep{deit}.  We computed the magnitude of element-wise difference, e.g. , $|W_Q - W_K|$,  and measured the percentage under a small threshold $\epsilon=1e-2$.  It is evident that the redundancy between $W_Q$ and $W_K$ is significantly higher than that between the other two pairs, as a result of the entanglement. 

\begin{figure}
  	\centering
  	\includegraphics[width=.8\textwidth]{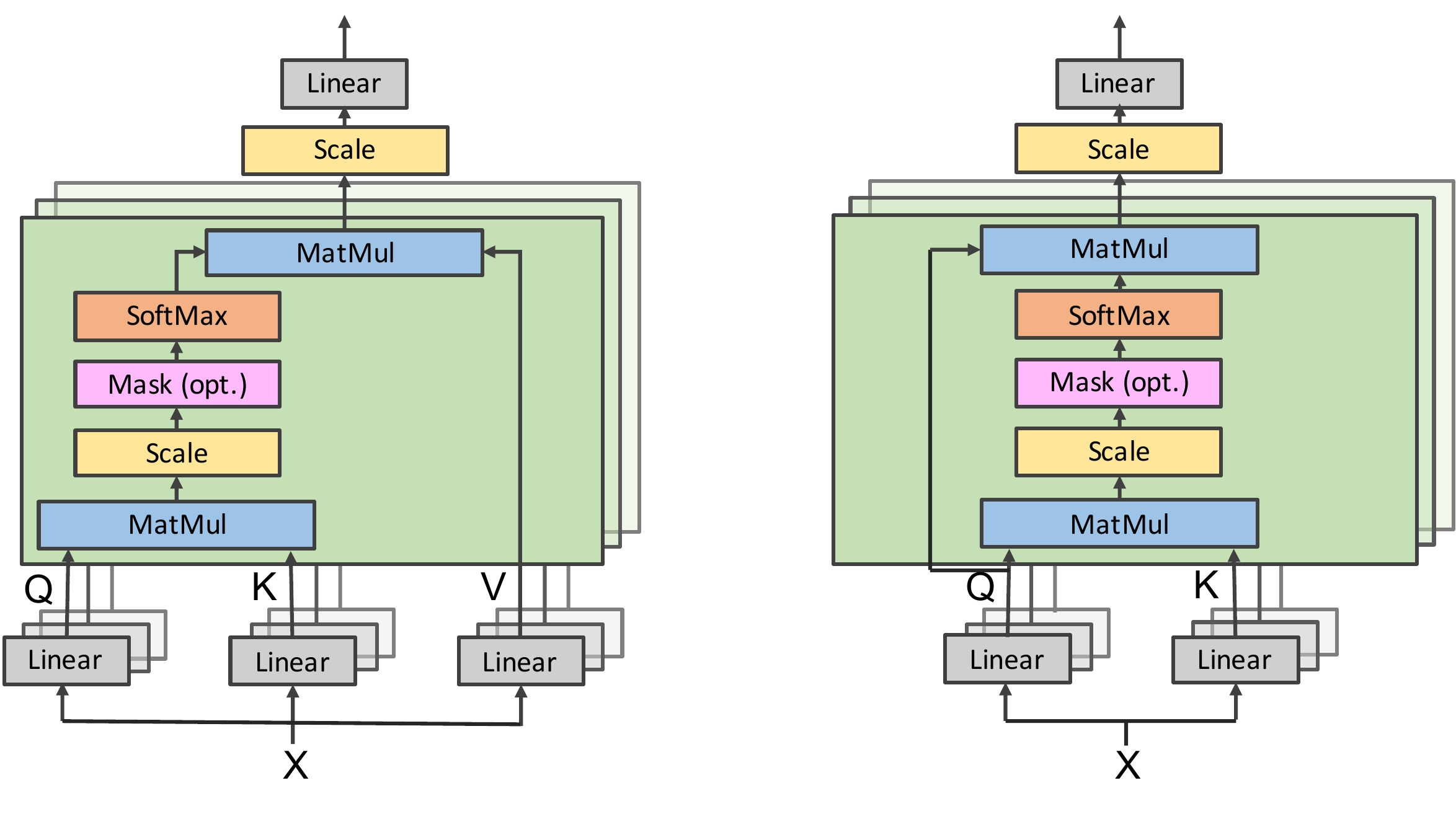}
  	\caption{Regular multi-head self-attention and Armour self-attention.}
  	\label{fig:mha}
\end{figure}

To exploit this fundamental redundancy,  we propose the \textit{Armour} self-attention by sharing the weights between $Q$ and $V$.  Therefore, its scaled dot-product attention becomes:
\begin{equation}
\textrm{Attention}(Q,  K) = softmax(QK^T / \sqrt{d}) Q.
\end{equation}
A comparison between the regular self-attention and the Armour variant is illustrated in Figure \ref{fig:mha}. The proposed method reduces the number of linear transformations from 3 to 2,  yielding smaller and faster transformers.  More importantly, by using $W_Q$ both before and after the softmax function,  we break the entanglement between $W_Q$ and $W_K$, forcing both to learn more efficient mappings.  Since Armour exploits the fundamental redundancy in attention,  it leads to optimized models with the same or better accuracies.

There are two other ways of weight sharing between $Q$, $K$ and $V$: $W_Q=W_K$ and $W_K=W_V$.  Intuitively,  $W_Q=W_K$ seems more natural, since $QK^T$ is where the entanglement happens.  In fact, this was \textit{accidentally} discovered in Reformer \citep{reformer} as a by-product of the locality-sensitive hashing (LSH) attention which requires $Q$ and $K$ to be equal.  However, the authors realized in a $W_Q=W_K$ formulation,  the dot-product between a query vector and itself will almost always be greater than the dot-product with another vector, which undermines the capacity of self-attention.  As a result, Reformer had to modify the masking to forbid a token from attention to itself,  except in situations where a token has no other valid attention targets.  This necessary but overly-simplified additional step incurs latency overhead during training and inference, and still reduces the capacity of the attention matrix along the diagonal.  The other variant is $W_K=W_V$, which is sometimes used in auto-regressive transformers where $Q$ is calculated from a different sequence than $K$ and $V$.  In this case, as shown in Equation \ref{eqn:dot_prod_atten},  an extra tensor transpose is required between $K^T$ and $V=K$, which will lead to performance penalty.  

In the first 4 rows of Table \ref{tab:deit_armour},  we compared the Armour attention against the other variants denoted as $W_K=W_V$, $W_Q=W_K$ diag, and $W_Q=W_K$.  Among all cases,  Armour yields the best ImagetNet accuracy and throughput.  The two variants of $W_Q=W_K$ with and without the diagonal post-processing both result in noticeable accuracy drops.  The Armour-Ti is produced by replacing the regular self-attention in the baseline DeiT-Ti \citep{deit} with the Armour attention.  The optimized model retains virtually the same accuracy while achieving a $7.8\%$ decrease in model size and a $7.2\%$ increase in GPU throughput.  It is also worth noting that the new best compact ViT, Armour-Ti dist/ 1200, is now in range with EfficientNet-B0 in terms of model size and accuracy, and offers a slight advantage in GPU throughput. This confirms the redundancy even in the smallest DeiT model,  and demonstrates Armour as a good replacement for the regular attention.  

\begin{table}
  \caption{DeiT vs.  Armour \& its variants}
  \label{tab:deit_armour}
  \centering
  \begin{tabular}{lrrr}
    \toprule
    Architecture     & \# params (M)     & INET Top-1 & GPU Throughput\\
    \midrule
    $W_K=W_V$  & 5.3 & 71.4\%    & 2,790  \\
    $W_Q=W_K$ diag  & 5.3 & 71.2\%    & 2,759  \\
    $W_Q=W_K$  & 5.3 & 70.9\%    & 2,791  \\
    Armour-Ti  & \textbf{5.3 }& 72.0\%    & \textbf{2,828}  \\
    DeiT-Ti & 5.7  & \textbf{72.1\%}  & 2,661  \\
    \midrule
    DeiT-Ti dist     & 5.9      & 73.8\%  & 2,649\\
    Armour-Ti dist     & 5.5       & 73.8\%  & 2,819\\
    Armour-Ti dist / 1200     & \textbf{5.5}       & 76.5\%  & \textbf{2,819}\\
    \midrule
    EfficientNet-B0 & 5.3      & 77.1\%  & 2,743\\
    \bottomrule
  \end{tabular}
\end{table}

\section{Generalized Compact Self-Attention}
\label{sec:generalized_compact_attention}
Since the Armour attention targets the fundamental redundancy in the regular self-attention.  It is highly generalizable and can be applied orthogonally on top of most ViT optimizations, as long as transformations of some form are still applied.  In this section,  we use a highly-competitive hybrid ViT architecture, LeViT \citep{levit},  as an example to show its drop-in applicability.  We derive two variants of the new attention design by applying the generalized Armour mechanism. For the other works that also modify the regular self-attention, their generalized compact variants can be similarly derived. 
\begin{figure}
  	\centering
  	\includegraphics[width=.98\textwidth]{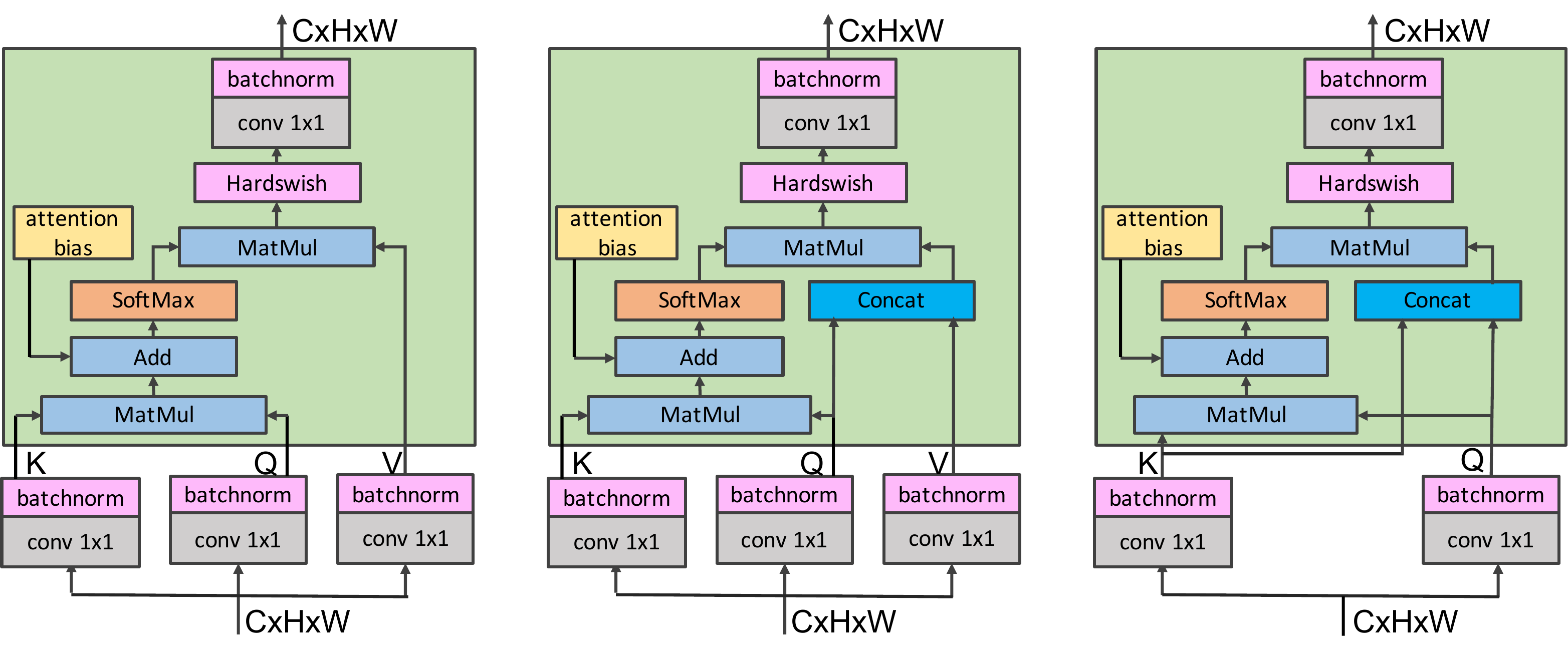}
  	\caption{The LeViT attention block and its compact variants. Left: regular attention block. Middle: with $V$ reduced by half and concatenated with $Q$ for the second matrix multiply. Right: with $V$ completed replaced by $Q$ and $K$. }
  	\label{fig:levit}
\end{figure}

There are two types of attention blocks in LeViT: the regular attention block depicted on the left in Figure \ref{fig:levit},  and the shrinking attention block. The latter accounts for only 2 out of 11 attention blocks in the smallest architecture, and is not considered in this work. The regular attention block uses $1\times1$ convolutions for the linear transformations, and replaces the LayerNorm with BatchNorm. In an effort to reduce the computation of $QK^T$, the size of $Q$ and $K$ matrices is reduced relative to $V$ by half. Therefore,  the shapes of $Q, K,$ and $V$ matrices are $N\times(D\times HW), N\times (HW\times D)$,  and $N\times (HW\times 2D)$, respectively, where $D$ is the key dimension, $N$ is the number of heads,  and $C\times H\times W$ is the traditional shape of activation maps in the $BCHW$ format.
 
Despite the changes in transformations and dimensions, we use this evolved attention block as an example to demonstrate how the Armour attention can be easily generalized to provide additional optimizations.  We derive two compact variants: one that reduces $V$ by half, and the other that removes $V$ completely.  The first variant is depicted in the middle of Figure \ref{fig:levit}, where the size of $V$ is changed to $N\times (HW\times D)$, and is \textit{concatenated} with $Q$ to maintain the $2\times$ ratio of $V$ in the original design.  The second variant is depicted on the right of Figure \ref{fig:levit},  where $V$ is completely replaced by the concatenation of $Q$ and $K$. 

In Table \ref{tab:levit_armour}, we show the results of applying the two variants to LeViT-128S: Armour-128S and Armour-128S+.  The first variant introduces a small reduction of the number of parameters by $3.7\%$,  and negligible improvement in GPU and CPU throughputs.  However,  we break the gradient entanglement between $W_Q$ and $W_K$ by concatenating $Q$ into $V$ and using $W_Q$ in both matrix multiplies.  As a result,  Armour-128S is $1.1\%$ more accurate than LeViT-128S, as all three 1$\times$1 convolutions are forced to learn better mappings.  The second variant offers more moderate improvements and a $0.6\%$ accuracy advantage compared to the baseline. 
\begin{table}
  \caption{LeViT 128S vs. Armour 128S/128S+}
  \label{tab:levit_armour}
  \centering
  \begin{tabular}{lrrrr}
    \toprule
    Architecture     & \# params (M)     & INET Top-1 & GPU Throughput & CPU Throughput\\
    \midrule
    LeViT-128S & 7.76  & 74.4\%                      & 12,880   & 77.6\\
    Armour-128S  & 7.47 & \textbf{75.5}\%    & 12,947    & 77.9\\
    Armour-128S+  & \textbf{7.18} & 75.0\%   & \textbf{13,035}  & \textbf{81.2}\\
    \bottomrule
  \end{tabular}
\end{table}

Note that we chose tensor concatenation as it is the simplest form of generalized Armour attention.  In the scenarios where $Q$, $K$, and $V$ are generated by convolutions with different shapes, the typical efficient convolution tricks, such as grouped convolution \citep{alexnet} and shuffled grouped convolution \citep{shufflenet} can also be considered.  With this example, we have demonstrated how easily the Armour attention can be generalized to exploit the redundancy even in the smallest model of a highly-optimized hybrid ViT architecture. 

\section{Experiments}
\label{sec:experiments}
In this section, we present experimental results demonstrating the techniques introduced above. The experiments are conducted with simple modifications on top of DeiT \citep{deit} and LeViT \citep{levit}, respectively. The ImageNet-2012 dataset is used for training and evaluation,  which is made possible for ViT by the extensive data augmentation techniques in \citep{deit}.  The training uses an 8x NVIDIA V100 GPU machine.  The inference time is measured on one V100 GPU and on an Intel Xeon E5-2686 at 2.3GHz.  The Armour models are trained with the same settings and hyper-parameters as their DeiT and LeViT counterparts, expect for a different learning rate of 1e-3 instead of 5e-4.

\subsection{Armour Attention Experiments}
In Table \ref{tab:deit_armour},  the first 5 architectures are trained without distillation for 300 epochs.  We were able to reproduce the reported accuracy for DeiT-Ti,  $72.1\%$ vs.  $72.2\%$.  Amour-Ti represents the same architecture of DeiT-Ti,  expect that all the regular self-attentions are replaced by the Armour attentions.  The optimized model achieves a $7.8\%$ decrease in the number of parameters and a $7.2\%$ increase in GPU throughput, while retaining virtually the same accuracy at $72.0\%$.  

The next 3 architectures in Table \ref{tab:deit_armour} are trained with distillation.  By training DeiT-Ti dist for 300 epochs, we observed an accuracy gap between the reproducible accuracy ($73.8\%$) and the published accuracy ($74.5\%$).  Despite that,  Armour-Ti dist achieves the same accuracy, while improving the model size by $6.8\%$ and  the GPU throughput by $6.4\%$.  Because of the accuracy reproduction issue, the most accurate Armour-Ti dist/ $1200$ is trained for $1200$ epochs instead of 1000 as in DeiT, which yielded a nearly identical accuracy as the best DeiT-Ti with distillation ($76.5\%$ vs.  $76.6\%$). This shows the Armour attention has indeed exploited the fundamental redundancy in the regular self-attention mechanism.

In Table \ref{tab:deit_armour_throughput}, we measured the improvements in model size and throughput for the other DeiT architectures.  Since we have verified the accuracy of the smallest optimized architecture in Table \ref{tab:deit_armour},  we expect these larger variants to safely benefit from the improvements without significant accuracy degradation.  

\begin{table}
  \caption{DeiT vs.  Armour: model size and throughput}
  \label{tab:deit_armour_throughput}
  \centering
  \begin{tabular}{lrr|rr}
    \toprule
    Architecture      & \# params (M) & \# params $\Delta$ & Throughput & Throughput $\Delta$ \\
    \midrule
    DeiT-Ti               & 5.7                 & -                          & 2661              & -\\
    Armour-Ti          & 5.3                 & -7.8\%                  & 2828              & +7.2\%\\
    \midrule
    DeiT-S               & 22.1                 & -                          & 987              & -\\
    Armour-S          & 20.3                 & -8.1\%                  & 1053              & +6.7\%\\
    \midrule
    DeiT-B               & 86.6                 & -                          & 313              & -\\
    Armour-B          & 79.5                 & -8.2\%                  & 335              & +6.9\%\\
	\bottomrule
  \end{tabular}
\end{table}

\subsection{Generalized Armour Attention Experiments}
In Table \ref{tab:levit_armour}, all three architectures are trained with distillation for 500 epochs. The best published accuracy of the baseline LeViT-128S is 76.6\%, which is produced by training with distillation for 1000 epochs.  Compared to the locally reproduced baseline, both Armour variants achieved better accuracies as well as slightly improved GPU and CPU throughputs.  In Table \ref{tab:levit_armour_throughput}, we also calculated the improvements in model size and throughput for the larger LeViT architectures, if the generalize Armour attention is applied.  Overall, the generalized Armour mechanism achieves good reduction of model size, and small to moderate throughput uplift. 

\begin{table}
  \caption{LeViT vs.  generalized Armour: model size and throughput}
  \label{tab:levit_armour_throughput}
  \centering
  \begin{tabular}{lrr|rr|rr}
    \toprule
    Architecture     & \makecell{\# params \\ (M)}    & $\Delta$ & \makecell{GPU \\ Throughput} & $\Delta$  & \makecell{CPU \\ Throughput} & $\Delta$ \\
    \midrule
    LeViT-192        & 10.9                   & -                & 8,565                  & -                & 37.1                        & -\\
    Armour-192     & 10.4                   & -4.6\%         & 8,695                   & +1.5\%          & 38.1                       & +2.7\%\\
    Armour-192+  &  9.8                    & -10.1\%        & 8,862                    & +3.5\%          &39.8                       & +7.3\%\\
    \midrule
    LeViT-256        & 18.9                   & -                & 6,557                 & -                & 23.5                       & -\\
    Armour-256     & 17.9                   & -5.3\%         & 6,675                   & +1.8\%          & 24.8                       & +5.5\%\\
    Armour-256+  &  17.0                    & -10.1\%        & 6,817                    & +4.0\%          &25.6                       & +8.9\%\\
    \midrule
    LeViT-384        & 39.1                   & -                & 4,136                   & -                & 12.1                        & -\\
    Armour-384     & 37.1                   & -5.1\%         &  4,226                   & +2.2\%          & 12.5                       & +3.3\%\\
    Armour-384+  & 35.0                   & -10.5\%      &  4, 314                   & +4.3\%          &13.0                       & +7.4\%\\
    \bottomrule
  \end{tabular}
\end{table}

\section{Conclusion}
\label{sec:conclusion}
In this paper, we re-examined the design of the widely-used self-attention and identified a fundamental redundancy.  To exploit that, we proposed \textit{Armour}, a compact self-attention that reduces the number of linear transformations.  Armour is easy to implement and highly generalizable.  By replacing the regular attention with Armour attention,  we produced a new state-of-the-art compact ViT that is in range with EfficientNet-B0 in model size and accuracy, with a slight advantage in throughput.  We have also demonstrated that the generalized Armour mechanism can be easily and orthogonally applied on top of the other  transformer optimizations, even when the design of self-attention has evolved.  Thanks to its drop-in applicability,  we believe Armour should be considered as a replacement for both regular attention and its variants for better accuracy, model size, and throughput.

\section*{Broader Impact}
The Transformer models have achieved remarkable success for many NLP tasks, and demonstrated promising potential for vision tasks.  However, these models typically require enormous computation and memory capacities, which limits their usage in settings with limited resources, such as mobile and embedded devices. Our work focuses on the fundamental redundancy in the self-attention mechanism that is the widely-used by transformers.  The proposed method can be easily generalized, therefore serving as ``a raising tide that lifts all boats".  By providing an easy-to-implement drop-in replacement, for self-attention,  numerous transformer models across many application domains can be seamlessly improved in model size, power and latency.  These optimized models can yield an immediate impact on real-life applications,  especially in settings with resource constraints.

\begin{ack}
We are grateful to Benjamin Graham,  Hugo Touvron,  and Francisco Massa for their generous help in reproducing the published results. 
\end{ack}

\bibliography{example_paper}
\bibliographystyle{plainnat}

\end{document}